\documentclass[a4paper,conference]{IEEEtran}
\usepackage[section]{placeins}
\usepackage{times}
\usepackage{float}
\usepackage{stfloats}
\usepackage{amsthm}
\usepackage{amsmath}
\usepackage{paralist}
\usepackage{url}
\usepackage{color}
\usepackage{framed}
\usepackage{mathrsfs}
\usepackage{graphicx}
\usepackage{bbm}
\usepackage{subfigure}
\usepackage{caption}
\usepackage{amsfonts}
\usepackage{multicol}

\usepackage{epstopdf}

\ifCLASSINFOpdf
\else
\fi
\begin{document}
\title{Anomaly Detection via Minimum Likelihood Generative Adversarial Networks}

 \author{\IEEEauthorblockN{Chu Wang$^{1}$ \qquad \qquad Yan-Ming Zhang$^{1}$ \qquad Cheng-Lin Liu$^{1,2,3}$}
\IEEEauthorblockA{$^{1}$National Laboratory of Pattern Recognition, Institute of Automation, Chinese Academy of Sciences \\
                        No.95 Zhongguancun East Road, Beijing 100190,  China\\
                  $^{2}$CAS Center for Excellence of Brain Science and Intelligence Technology, Beijing, China\\%, Institute of Automation, Chinese Academy of Sciences \\
                  $^{3}$University of Chinese Academy of Sciences, Beijing, China \\
	Email: $$chu.wang$$@ia.ac.cn ~~ $\{$ymzhang, liucl$\}$@nlpr.ia.ac.cn }
         }


%Telephone: (800) 555--1212, Fax: (888) 555--1212}


\maketitle


\begin{abstract}
 Anomaly detection aims to detect abnormal events by a model of normality. It plays an important role in many domains such as network intrusion detection,  criminal activity identity  and so on. With the rapidly growing size of accessible training data and high computation capacities, deep learning based anomaly detection has become more and more popular. In this paper, a new domain-based anomaly detection method based on generative adversarial networks (GAN) is proposed. Minimum likelihood regularization is proposed to make the  generator produce more anomalies and prevent it from converging to normal data distribution. Proper ensemble of anomaly scores is shown to improve the stability of discriminator effectively. The proposed method  has achieved significant improvement than other anomaly detection methods  on Cifar10 and UCI datasets.
\end{abstract}


% \fi
\IEEEpeerreviewmaketitle

\section{Introduction}
 Anomaly detection refers to the process of modeling normal events and detect  abnormal ones. It has been widely applied in many domains, such as  electronic IT security  which involves network intrusion detection and fraud detection. In \cite{jyothsna2011review}\cite{yang2006support}, anomaly detection is applied to detect possible intrusions such as  malicious activity, computer attack, computer misuse and virus spread.
In \cite{schreyer2017detection}, a method is proposed to  detect fraud in large-scale accounting data, which is also important in financial statement audits or forensic investigations. In \cite{schlegl2017unsupervised}, anomalies are identified in medical imaging data to capture imaging markers relevant for disease progression and treatment monitoring. Anomaly detection is also applied in industrial monitoring and damage detection \cite{clifton2006learning}, image processing and video surveillance \cite{pokrajac2007incremental}, text mining \cite{ando2007clustering} and  sensor network \cite{chatzigiannakis2006hierarchical} etc.
%Anomaly detection is also applied in industrial monitoring and damage detection, image processing and video surveillance, text mining, sensor network.

Existing approaches for anomaly detection can be divided into five categories: probability-based, distance-based, reconstruction-based, domain-based and information-theory-based. Probability-based approaches \cite{chatzigiannakis2006hierarchical}\cite{erdogmus2004multivariate} are based on generative probability density function of a given dataset.  They use only a small amount of information. But their performance  is limited in high dimensional space.
Distance-based approaches include clustering \cite{angelov2004approach}\cite{BDLZ2014} and nearest neighbour methods \cite{stranjak2008multi}\cite{DU2016115}. Such approaches depend on a well-defined metric to compute the distance between two data points. Distance-based approaches do not require to know the data distribution, but rely on a suitable distance metric to estimate the similarity between two data points. They are not flexible enough to detect local anomalies that have diverse densities and arbitrary shapes.
Domain-based  approaches aim to build a boundary of normal data. Support vector data description (SVDD) \cite{campbell2001linear} and one-class SVM (OCSVM) \cite{heller2003one} are two instances of these approaches. Domain-based methods are insensitive to sampling and the density of target class. Information-theory-based approaches assume that anomalies significantly alter the information of a dataset.  They aim to find points whose elimination from the dataset induce the biggest difference of information. These approaches  make no assumptions about the distribution of a given dataset, but work well only if there is a significantly large number of anomalies. Reconstruction-based approaches are mostly neural-network-based  \cite{augusteijn2002neural} and subspaces-based \cite{mcbain2011feature}. In neural-network-based approaches, the deviation between  target value and the output of  neural network is used to measure the anomalies.

The above methods for anomaly detection have  good  mathematical basis, but their performance is limited by the effectiveness of feature extraction. Deep neural networks can overcome such drawbacks.  As well known deep generative neural network, autoencoders (AE) and variational autoencoders (VAE) have been widely used for anomaly detection. Data points which have large reconstruction errors or reconstruction probabilities are regarded as anomalies. In \cite{chalapathy2017robust}, robust autoencoders  which capture the majority of data points while allowing for some data to have arbitrary corruption has shown better performance on anomaly detection.   There are also some work \cite{principi2017acoustic}\cite{ravanbakhsh2017abnormal}\cite{schlegl2017unsupervised} for anomaly detection based on GAN. In all of these methods, the generators of GAN are trained to produce samples and fit the data distribution. In the testing phase, the anomaly score of a test data $x$ is computed by evaluating the probability of generating $x$ with the learned generator. Therefore, such methods belong to the category of probability-based methods.

%Unsupervised learning problem is closely connected with the following problems.

%In incremental learning, raw data that come from environment  become incrementally available for a trained model. As a branch of incremental learning, concept drift aims to detect changes in data stream. Such change occurs not only because of anomalies in data stream but also  because of non-stationary environments. %In the second case, the data from new environment are collected to form a new model.

%Anomaly detection is closely connected to many important problems in machine learning. Multi-class open set classification \cite{bendale2016towards}\cite{mandelbaum2017distance}\cite{yu2017open} is trained not only to classify the data points belonging to  known classes, but also identify whether a test data belongs to known classes or not.  The method of \cite{ge2017generative} extends that of \cite{bendale2016towards} by training open-max classifier with anomalies selected from the generator of a pretrained GAN. Class discovery aims to identify a subset of training set belonging to new classes beyond known classes. Compared with anomaly detection, this subset may consist of  rare classes, not only anomalies. Zero-shot learning  aims to transfer knowledge from seen classes to unseen classes which do not appear in training set. Novel approaches were proposed in  \cite{palatucci2009zero}\cite{socher2013zero} to describe the information about unseen classes. Compared with anomaly detection, zero-shot learning assumes that some  information about unseen classes is available during transferring.

In this paper, we propose a GAN-based anomaly detection method. Different from previous methods, our method uses the discriminator of GAN to detect anomalies, and thus belongs to the category of domain-based anomaly detection methods. The core idea is that we use both normal data and the anomalies produced by the generator of GAN to train a discriminator in the hope that the boundary of normal data can be correctly captured by the discriminator. Minimum likelihood regularization is developed to make the generator produce more anomalies during training and prevent the generator from converging to normal data distribution. Furthermore, we adopt ensemble learning to overcome the instability of GAN. We compare our method with other anomaly detection methods including OCSVM, IFOREST, VAE, AE. The experimental results show that our method achieves better performance on Cifar10 and several UCI datasets.

The remaining  of the paper is organized as follows: Section 2 reviews related work. Section 3 introduces basic knowledge of variational inference and GAN, on which our method is based. Section 4 introduces the proposed method with minimum likelihood  regularization and ensemble learning to improve the performance of discriminator on anomaly detection. Section 5 presents the experimental results and  Section 6 concludes the paper.

\section{Related Work}
%We propose an one-class classification method based on GAN for  domain-based anomaly detection.  We set the output  before sigmoid activation function  in the discriminator of  to be $l(x)$ and the anomaly score to be $-l(x)$. In the following, we show  GAN can be used as an one-class classifier.
In this section, we review several domain-based anomaly detection methods and  GAN-based anomaly detection methods. Some representative methods are outlined below.

%In this section, we first review several domain-based methods and then introduce existing anomaly detection methods based on GAN. As an important branch of anomaly detection methods, domain-based methods aim to describe the boundary of normal data in feature space. There are many conventional domain-based methods for anomaly detection such as OCSVM, SVDD and so on. We briefly describe some representative methods and discuss the difference of between these methods and ours.

OCSVM: OCSVM  proposed in \cite{scholkopf2001} is a well-known anomaly detection method. In OVSVM, the origin in the feature space is set to be the only anomaly data. A hyperplane is trained to distinguish normal data and the origin in feature space. The signed distance between a given point and such boundary is defined as anomaly score. The feature space is constructed by a given kernel function.

SVDD: In $\cite{campbell2001linear}$, SVDD is proposed to boundary the normal data in feature space by a sphere. The sphere is optimized to contain normal data with the smallest volume. In this method, anomalies are  unnecessary for the construction of the  boundary of normal data.

ASG-SVM: In \cite{yu2017open}, one way of generating normal data and  anomalies in unsupervised manner via an adversarial learning strategy is proposed. Such generated points are trained by a discriminator to form the boundary of normal data.  The way of producing normal data and anomalies is different from our method.

VAE/GAN: In \cite{larsen2015autoencoding},  VAE is trained  by the discriminator of  GAN instead of element-wise reconstruction objective. Such model and its variants have been widely used for anomaly detection. Given a test point, the anomaly score is  defined according to the distance to the reconstructed one and the distance is measured by the discriminator of GAN instead of $L_2$ loss function. In our method, anomaly score is defined according to the output of the discriminator in GAN. The output of generator is not directly related to the anomaly score.

AnoGAN: In \cite{schlegl2017unsupervised},  convex combination of two distances is computed through GAN as the anomaly score of a  test point $x$. The first one is  $\|x-g_\theta(z)\|$ where $g_\theta(z)$ is the output of generator under $z$. The second one is  $\|d(x) -d(g_\theta(z))\|$ where $d(x)$ is the output of discriminator's intermediate layer. The anomaly score measures the distance between $x$ and its nearest point $z$ produced by the generator of GAN. While in our method, anomaly score is used to measure the distance between $x$ and the boundary of normal data defined by the output of the discriminator in GAN.

% In \cite{}, a robust version of convolutional autoencoder is proposed to achieve better performance. In \cite{an2015variational}, the reconstruction probabilities of variational autoencoder(VAE) is shown to have a better performance than the reconstruction errors for classical autoencoder. Combined with other network structures,  a lof of variants of  autoencoders are developed for  more tasks and more types of data. \cite{principi2017acoustic} proposes adversarial autoencoders in \cite{makhzani2015} to train on a corpus of normal acoustic signals and  detect whether a segment contains an abnormal event or not.  Recurrent autoencoders in \cite{} can model the probability density function of time series. %\cite{hawkins2002outlier, bontemps2016collective, malhotra2016lstm} proposes recurrent autoencoders to detect the anomalies on time series dataset.

Bad-GAN: In \cite{dai2017good}, KL-divergence is used to disturb the generator of GAN to achieve better performance on semi-supervised problem. Although the motivation is the same as ours, the construction of KL-divergence is different from ours. KL-divergence in \cite{dai2017good} aims to minimize the distance between the probability induced by generator  and a constructed distribution.  In this paper, we use KL-divergence  to make the generator have low probability values (i.e., minimize the likelihood) on normal data via variational inference. We show that minimum likelihood regularization is  effective in anomaly detection.

\section{Background Methods}
Our method closely involves existing techniques variational inference and GAN, which are outlined in the following.

\subsection{Variational Inference}

Let $p_{data}(x| \theta)$ be a data distribution where $\theta$ is the parameter. Maximum likelihood estimation (MLE)  is  a classical method to fit $p_{data}(x|\theta)$. %When $p_{data}(x|\theta)$ is complex, computing MLE of $p_{data}(x|\theta)$ directly is difficult.
When $p_{data}(x|\theta)$ is complex, it can be approximately estimated by variational inference. Variational inference aims to optimize the evidence lower bound $\mathcal{L}(x, \theta, q)$  defined as follows:
\begin{align*}
\mathcal{L}(x, \theta, q):&= \log p_{data}(x| \theta) - D_{KL}(q(z|x)\| p(z|x;\theta))\\
&= \int q(z|x)\log p(x|z, \theta) - D_{KL}(q(z|x)\| p(z)),
\end{align*}
where $q(z|x)$ represents a conditional distribution and $D_{KL}(\cdot\|\cdot)$ is denoted as KL divergence. It is well-known that $\log p_{data}(x|\theta)= \max\limits_{q}\mathcal{L}(x, \theta, q)$ and the maximum is attained if and only if $q(z|x) = p(z|x,\theta)$.
\subsection{Generative Adversarial Networks}
Generative adversarial networks (GAN) proposed in \cite{NIPS2014} is a widely used deep generative model. A lof of variants such as \cite{eghbal2017likelihood}\cite{grover2017flow} have been developed for improving the performance. The basic idea of GAN is to  train a generator $G$ and a discriminator $D$ such that $D$ learns to distinguish whether a sample is real or fake and $G$ learns to fool  discriminator $D$. The objective function of GAN is the following minmax game:
\begin{equation*}%\label{GAN}
\min\limits_{G}\max\limits_{D}\mathbb{E}_{x\sim p_{data}(x)}\log D(x) + \mathbb{E}_{z\sim p(z)}\log (1-D(G(z))).
\end{equation*}
%The objective function $G$ under optimal discriminator $D$ is Jensen-Shannon divergence. Global minimum of $C(G)$ is achieved when  $p_{data}=p_G$, where $p_G$ is induced by $G$.

In \cite{salimans2016}, feature matching is designed to prevent generator $G$ from overtraining on  discriminator $D$.  Let $f(x)$ be the activations on an intermediate layer of discriminator $D$, the objective of generator $G$ is defined as \[\| \mathbb{E}_{x\sim p_{data}}f(x) - \mathbb{E}_{z\sim p(z)}f(G(z))\|.\] Feature matching aims to match the first moment of $p_{data}$ and the distribution $p_G$ induced by $G$, but not $p_{data}$ and $p_G$ themselves. Although $p_{data}$ is a fixed point of $G$, it is not necessary that $p_G$ converges to $p_{data}$ during training.

\section{Proposed Method}
   Our method detects anomalies using the discriminator of GAN while producing  abnormal data using the generator of GAN. In initial phase of  training GAN, the generator outputs nearly random samples which are regarded as weak anomalies compared with normal data. In this case, the discriminator $D$ is trained to have high values on normal data and low values on such random samples. The boundary between normal data and anomalies defined by the output of  discriminator $D$ is far from normal data. As the  outputs of generator approach to the  normal data during training, such boundary becomes compacter to normal data and form the boundary of normal data at last.

Although  discriminator $D$  can detect anomalies during training as shown above, it also meets some troubles during training. This is because the induced probability $p_G$ by generator $G$  converges to normal data distribution $p_{data}$ and discriminator $D$ converges to $\frac{1}{2}$ when it has enough capacities. In this case, the performance of $D$ will degenerate in final phase of training.

In the following, we propose a novel regularization method for the generator of GAN to achieve better performance on anomaly detection. Furthermore, ensembling learning is used to overcome the instability of GAN in our method.

\subsection{Minimum Likelihood GAN}
 To deal with the degeneration of discriminator $D$ during training and improve the performance of $D$ for anomaly detection, we regularize $G$ such that
\begin{itemize}
\item $G$  produces more anomalies during training.
\item $p_G$  does not converge to $p_{data}$.
\end{itemize}

To achieve this goal,  KL divergence is proposed to prevent $p_G$ from converging to $p_{data}$. Let $z\sim p(z)$ where $p(z)$ is the prior distribution of generator $G$. Since $p_G$ is the distribution of  $G(z)$, the support of $p_G$ is usually a manifold in high dimensional space. In this case, $KL(p_{data}\|p_G)$ is not well defined. Define random variable $x$  as  $x:= G(z) + n$ where $n$ is an independent random variable from $z$. The distribution of $n$ can be Gaussian distribution or Laplace distribution. Define $\tilde{p}_G$ to be the distribution of $x$. Since \[p(x|z)>0, \quad p(z)>0\]  for each $x$ and $z$, we have that \[\tilde{p}_G(x) = \int p(x|z)p(z)dz>0\] for each $x$. Therefore, the support of  $\tilde{p}_G$   is the whole space. Furthermore, $\tilde{p}_G\approx p_G$, when $n$ is properly chosen.  The objective function of  $G$ is defined as follows:
\begin{equation*}
\| \mathbb{E}_{x\sim p_{data}}f(x) - \mathbb{E}_{z\sim p(z)}f(G(z))\|- aKL(p_{data}\|\tilde{p}_G).
\end{equation*}
Minimizing $-KL(p_{data}\|\tilde{p}_G)$ is equivalent that $\tilde{p}_G$ has low values on normal data. We call GAN with such regularization  Minimum Likelihood GAN (MinLGAN).

Since $KL(p_{data}\|\tilde{p}_G) = \int p_{data} \log p_{data}-\int p_{data}\log \tilde{p}_G$
and $\tilde{p}_G$ has no close form, it is intractable  to compute the gradient of  $KL(p_{data}\|\tilde{p}_G)$ directly. In this case, we replace $\log\tilde{p}_G$ by $\max\limits_{\vartheta}\mathcal{L}(x, \theta, q(z|x, \vartheta))$ via variational inference.
Then $KL(p_{data}\|\tilde{p}_G)$ is approximated as follows:
\[\int p_{data} \log p_{data}-\int p_{data}\max\limits_{\vartheta}\mathcal{L}(x, \theta, q(z|x, \vartheta)).\]

Our algorithm consists of updating the parameters of  discriminator $D$, $q(z|x, \vartheta)$ and generator $G$ iteratively. For discriminator $D$, the objective function is
\[\max\limits_{D}\mathbb{E}_{x\sim p_{data}}\log D(x) + \mathbb{E}_{z\sim p(z)}\log (1-D(G(z))).\]
For $q(z|x, \vartheta)$, the objective function is
\[\max\limits_{\vartheta} \int q(z|x, \vartheta)\log p(x|z, \theta) - D_{KL}(q(z|x, \vartheta)\| p(z)).\]
For generator $G$, the objective function is
\begin{footnotesize}
\[\min\limits_{G}\| \mathbb{E}_{x\sim p_{data}}f(x) - \mathbb{E}_{z\sim p(z)}f(G(z))\| + a\int q(z|x, \vartheta)\log p(x|z, \theta).\]
\end{footnotesize}
There is a
geometric intuition  for variational inference when $q(z|x, \vartheta)$ and $p(x|z)$ are Gaussian or Laplace distribution. In the objective function of $q(z|x, \vartheta)$, maximizing  $\int q(z|x, \vartheta)\log p(x|z, \theta)$ means that $q(z|x, \vartheta)$ is trained to find $z$ such that $G(z)$ is close to $x$. In the objective function of $G$, minimizing $\int q(z|x, \vartheta)\log p(x|z, \theta)$  means that  $G(z)$ is trained to get away from $x$. In this case, KL regularization prevents  $G$ from generating normal data.

\subsection{Overcoming the Instability of Discriminator}
%Although a lot of methods have been developed to train GAN  in a more stable way, most of such methods aims to improve the quality of pictures generated by $G$. In anomaly detection, we focus on the performance of discriminator instead of generator and we have to treat the instability of discriminator.
%
%At first, we show that the all anomalies generated by the generator from the beginning of training GAN have an effect on the qualities of the discriminator to identify the anomalies.

The performance of discriminator $D$ depends on the trajectories of anomalies produced by generator $G$ during training. But trajectories of anomalies meet with randomness and uncertainties during training, such as random initial weights values and random sampling from prior distribution. Such randomness causes some instabilities of discriminator $D$ during training.

Ensemble learning is an effective way to deal with such instabilities. It combines two or more base learners to reduce bias and variance effectively. Two commonly used ensemble learning methods are bagging and boosting. Bagging involves multiple models in the ensemble which are obtained by using randomly drawn subsets of the training set, while in boosting they are achieved by emphasizing the training instances that previous models misclassify.

Similar to bagging and boosting, we train $N$ discriminators $D_i$ independently and compute $D_i(x)$ for each test sample $x$. For numeric stability, $D_i(x)$ refers to the output before sigmoid activation function in the last layer of discriminator $D_i$.  Anomaly score  $s$ of $x$ can be computed in two  ways. One way called ensemble GAN is defined as follows: \[s = -\frac{1}{N}\sum D_i(x).\]  Assume a holdout set $S$ is available.  Let $m_i:=\max\limits_{x\in S} D_i(x)$ and $n_i:=\min\limits_{x\in S} D_i(x)$. The second  way called scaled ensemble GAN is defined as follows:
\[s = -\frac{1}{N}\sum(D_i(x)-n_i) / (m_i-n_i).\]

\section{Experimental Results}

In this section, we first visualize the effects of KL regularization on circle and moon toy datasets, and then present experimental results of our  method on Cifar10 and UCI datasets. We show that our method achieves better performance than existing methods on such datasets.

\subsection{Visualizations on Toy Datasets}
\begin{figure}[h]
\center
\includegraphics[width=0.4\textwidth]{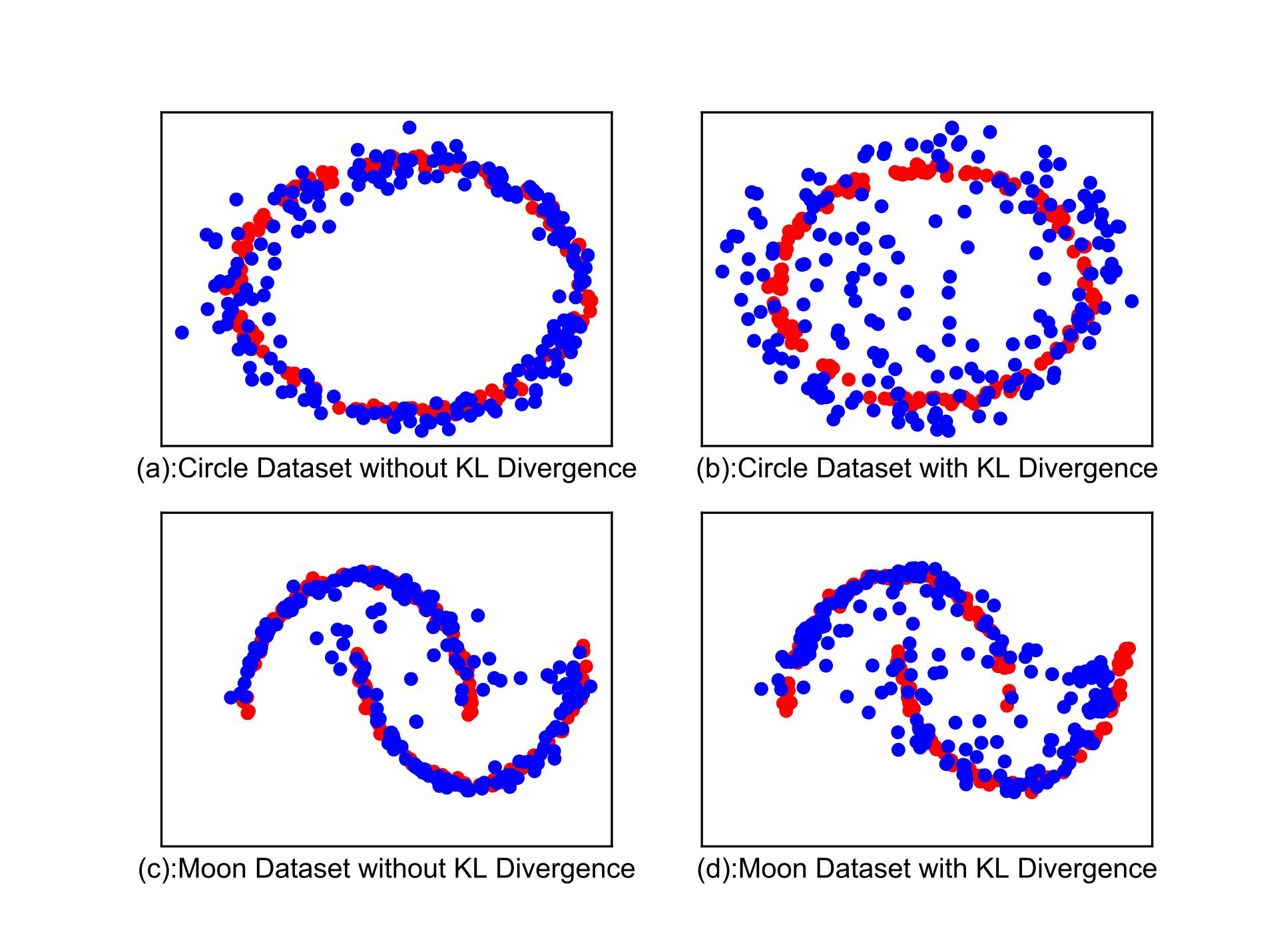}
\setlength{\belowcaptionskip}{-0.3cm}
\caption{\label{visualize}The performance of generator $G$ under KL divergence. The red points  are generated by  normal data distribution and the blue points are generated by $G$. Figure (a) and Figure (c) represent the performance without KL divergence. Figure (b) and Figure (d) represent the performance with KL divergence.}
\end{figure}

We select circle and moon toy datasets to visualize the performance of KL divergence on generator $G$.  In Figure \ref{visualize}(a) and \ref{visualize}(c), most of the blue points lie in normal data manifold. This shows that the distribution induced by $G$ is nearly the same as normal data distribution.   In Figure \ref{visualize}(b) and \ref{visualize}(d), many blue points lie outside the manifold. Since the coefficient of KL divergence is relatively small, most of blue points lie in a small neighborhood of data manifold. These two experiments show that KL-divergence is effective in making generator $G$ produce more points near normal data manifold and prevent $G$ from converging to normal data distribution.
%\begin{figure}[h]
%
%\center
%\subfigure[\label{circle-origin}]{
%\includegraphics[width=0.22\textwidth]{figure1.jpg}
%}
%\hfill
%\subfigure[\label{circle-kl}]{
%\includegraphics[width=0.22\textwidth]{figure2.jpg}
%}
%\vfill
%\subfigure[\label{moon-origin}]{
%\includegraphics[width=0.22\textwidth]{figure3.jpg}
%}
%\hfill
%\subfigure[\label{moon-kl}]{
%\includegraphics[width=0.22\textwidth]{figure4.jpg}
%}
%\caption{Visualization on Toy Datasets.}
%%\subfigure[Mnist  Anomaly Score\label{circle-dec}]{
%%%\includegraphics[width=1\textwidth]{mnist_prc.jpg}
%%\includegraphics[width=0.2\textwidth]{twomoon.jpg}
%%}
%\end{figure}

%\subsection{Experimental setup on benchmark datasets}
 %MNIST data set contains $60000$ training and $10000$ test handwritten digits images. The  size of images  is $28\times 28$ and the digits are from $0$ to $9$.
 %Three UCI datasets includes KDDcup 99, shuttle and forest cover. KDD dataset consists of five main classes. They are Normal class, DoS class, Probe class, R2L class and U2R class.  Only Normal class is normal data. All the other classes are called anomalies. KDD dataset consists of both discrete and continuous attributes. The scales of the continuous attributes  are different from each other. The shuttle dataset contains 9 attributes all of which are numerical. Approximately 80\% of the data belongs to class 1. Cover type Dataset predicts $7$ forest cover types from cartographic variables. It consists of $581012$ vectors and each vector consists of 54-dimensional attributes.
%\begin{table}[tbp]
%\centering\small
%\caption{\footnotesize KDD dataset\label{kdd}}
%\begin{tabular}{c|c|c|c|c|c|c|c|c|c|c|c|c}
%{p{1cm}|p{1cm}|p{1cm}|p{1cm}|p{1cm}|p{1cm}|p{1cm}}
%\hline
%{class} & {number of instances}\\
%\hline

% normal&   972,781  \\
%\hline
% dos&     3,883,370\\
%\hline
% Probe&  41,102   \\
%\hline
% r2l&    1,126 \\
%\hline
% u2r&    52 \\
%\end{tabular}
%\end{table}

\subsection{Experimental Results on Benchmark Datasets}
In these experiments,  training set only consists of normal data. The performance is evaluated on test data which contains both normal data and anomalies.  Our methods include Minimum Likelihood GAN (MinLGAN), ensemble MinLGAN (EMinLGAN-1) and scaled  ensemble MinLGAN (EMinLGAN-2). We compare our  methods with GAN baseline (GAN), OCSVM \cite{scholkopf2001}, IFOREST \cite{liu2008isolation}, VAE, AE. We implement our methods by Theano and our code is based on the code in https://github.com/openai/improved-gan. OCSVM  and  IFOREST are implemented by LIBSVM software \cite{chang2011libsvm}. Anomaly scores for such methods are defined as follows:

\begin{itemize}
\item Anomaly score for GAN is defined as the negative of discriminator's output.
\item Anomaly score for OCSVM is defined as   signed distance to  decision boundary.
\item Anomaly score for IFOREST is defined according to the number of splitting required to isolate a sample.
\item Anomaly score for AE is defined with respect to the reconstruction error.
\item Anomaly score for VAE is defined with respect to the reconstruction probability.
\end{itemize}

In our experiments, all anomalies are denoted as positive class and  normal data are denoted as negative class.  ROC curve is used to measure the performance of our method. The performance of different methods are measured by ROC scores.
\begin{itemize}
\item ROC  curve: plotting the true positive rate (TPR) against the false positive rate (FPR) at various threshold settings.
\item ROC score: the area under the ROC curve.
\end{itemize}

\subsubsection{Experimental Results on Cifar10 Dataset}
Cifar10 dataset consists of 60000 32x32 color images in 10 classes, with 6000 images per class. There are 50000 training images and 10000 test images.
We make 10 experiments where each of 10 classes is regarded as normal data and others are anomalies. All the experiments share the same network structure,  learning rate and regularization coefficient $a$. A small holdout set is used to decide the termination for each method. We repeat 80 times for each experiment and record the best performance on such holdout set for each time. The averaged ROC scores are shown in Table \ref{cifar10table}.

%\begin{table}[h]
%\caption{AUC ROC on Cifar10 Dataset\label{cifar10table}}
%\center
%\small
%\begin{tabular}{ccccccccccccccc}
%%{p{1cm}|p{1cm}|p{1cm}|p{1cm}|p{1cm}|p{1cm}|p{1cm}}
%\hline
%{Normal}& {MMDGAN}&{MDGAN}& {DGAN} & {ADGAN} & {AnoGAN} & {OCSVM} & {VAE} & {AE}\\
%\hline
%
% 0& 0.803&0.773&0.73   & 0.632& 0.610&  0.666& 0.620 &0.656  \\
%\hline
% 1& 0.601&0.594&0.596   & 0.529& 0.565& 0.473& 0.664 &0.435  \\
%\hline
% 2& 0.669&0.638&0.634   & 0.580& 0.648& 0.675& 0.382 &0.381  \\
%\hline
% 3& 0.598&0.578&0.569   & 0.606& 0.528& 0.530& 0.586 &0.545  \\
%\hline
% 4& 0.716&0.670&0.643   & 0.607& 0.670& 0.827& 0.386 &0.288  \\
%\hline
% 5& 0.684&0.594&0.6   & 0.659& 0.592& 0.428& 0.586 &0.643  \\
%\hline
% 6& 0.726&0.679&0.728   & 0.611& 0.625& 0.787& 0.565 &0.509  \\
%\hline
% 7& 0.626&0.603&0.546   & 0.630& 0.576& 0.532& 0.622 &0.690  \\
%\hline
% 8& 0.785&0.742&0.701   & 0.744& 0.723& 0.720& 0.663 &0.698  \\
%\hline
% 9& 0.608&0.583&0.583   & 0.644& 0.582& 0.453& 0.737 &0.705  \\
%\hline
% &  0.681&0.645&0.633  & 0.624& 0.612& 0.610& 0.581 &0.583  \\
%\end{tabular}
%\end{table}

\begin{table*}[t]
\setlength{\abovecaptionskip}{0pt}
\setlength{\belowcaptionskip}{20pt}
\caption{ROC scores on Cifar10 dataset.\label{cifar10table}}
\center
\begin{tabular}{ccccccccccccccc}
%{p{1cm}|p{1cm}|p{1cm}|p{1cm}|p{1cm}|p{1cm}|p{1cm}}
\hline
{Normal}&{EMinLGAN-1}& {EMinLGAN-2}&{MinLGAN} &{GAN}&{IFOREST} &{OCSVM} & {VAE} & {AE}\\
\hline

 0& 0.814&0.821&0.786&  0.76&  0.615&  0.689& 0.645 &0.739  \\
\hline
 1& 0.633& 0.642&0.61&  0.627& 0.688 &  0.464& 0.519 &0.358  \\
\hline
 2& 0.660& 0.664&0.643& 0.635& 0.476&  0.679& 0.638 &0.692  \\
\hline
 3& 0.568& 0.585&0.567& 0.589& 0.538&    0.513& 0.539 &0.575  \\
\hline
 4& 0.702&0.701&0.676& 0.664& 0.661&    0.767& 0.771 &0.774  \\
\hline
 5& 0.643&0.672&0.621&0.6&  0.607& 0.529& 0.505 &0.59  \\
\hline
 6& 0.732&0.721&0.697&0.706& 0.757 &  0.765& 0.715 &0.699  \\
\hline
 7& 0.623&0.62&0.599&0.565&  0.659&   0.53& 0.506 &0.515  \\
\hline
 8& 0.771&0.788&0.755&0.715&  0.7&   0.706& 0.73 &0.792  \\
\hline
 9& 0.639&0.652&0.616&0.604&  0.711&  0.481& 0.605 &0.42  \\
\hline
 Average&  0.679 &0.687&0.657&0.647& 0.641&  0.613& 0.617 &0.615 \\
\hline
\end{tabular}

\end{table*}
\begin{figure}[h]
\center
\includegraphics[width=0.33\textwidth]{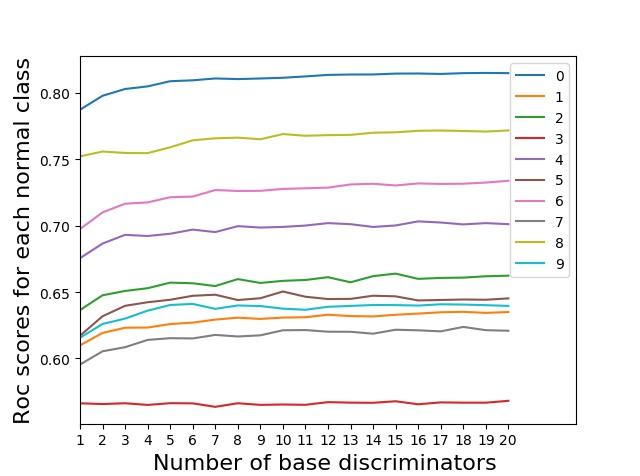}
\caption{ Ensemble ROC scores on Cifar10 dataset\label{cifar-curve}. The ensembled performance improves as the number of base discriminators increases and become stable when such number is more than 10.}
\setlength{\belowcaptionskip}{-0.2cm}
\end{figure}
\begin{figure}[h]
\center
\includegraphics[width=0.35\textwidth]{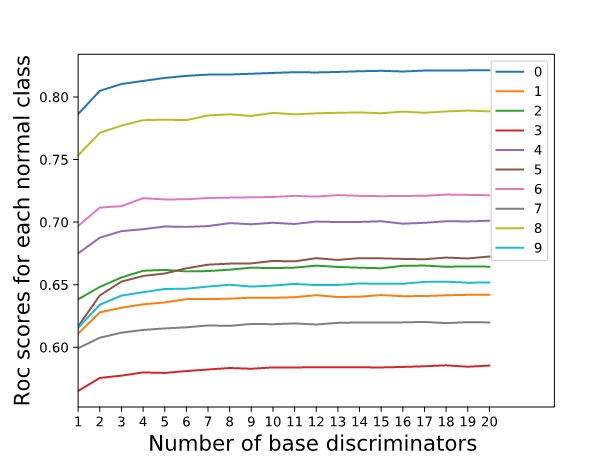}
\setlength{\belowcaptionskip}{-0.3cm}
\caption{Scaled ensemble ROC scores on Cifar10 dataset\label{cifar-distribution}. The ensembled performance improves as the number of base discriminators increases and become stable when such number is more than 5.}

\end{figure}

 Table \ref{cifar10table} shows that MinLGAN has better performance than GAN on class $0, 2, 4, 5, 7, 8, 9$. For class 1,3,6, MinLGAN is worse than GAN. This is because we use the same learning rate and regularization $a$ for each experiment. Although Kl regularization can help to produce more anomalies during training, if $a$ is too large, dynamics of GAN will be damaged. For class 1 and 3, both MinLGAN and GAN perform badly.  This is because that  the performance of  GAN is  sensitive to the network structure than other methods. Modifying neural network structure can improve the performance on such two classes. The performances of some discriminators are much lower than the averaged performance because of the instabilities of GAN.  Such ROC scores are also included in our results.

Figure \ref{cifar-curve} and Figure \ref{cifar-distribution} show the relationship between ROC scores and the number of base discriminators. We see that ROC scores become stable when the number of base discriminators is 5 for EMinLGAN-2, but 10 for EMinLGAN-1. The convergence rate of EMinLGAN-2 is higher than EMinLGAN-1. When both methods converge, EMinLGAN-2 performs better than EMinLGAN-1. This is because the order of the anomaly scores for each base discriminator is quite different. Anomaly scores produced by EMinLGAN-1 depend heavily on  base discriminators whose anomaly scores are in large order.

\begin{figure}[h]
\small\center
\includegraphics[width=0.33\textwidth]{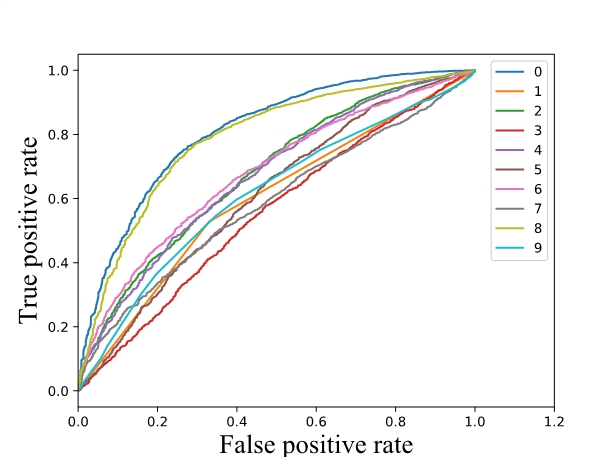}
\caption{ROC curves on Cifar10 dataset\label{cifar-roc-curve}. }
\end{figure}
\begin{figure}[h]
\small\center
\includegraphics[width=0.35\textwidth]{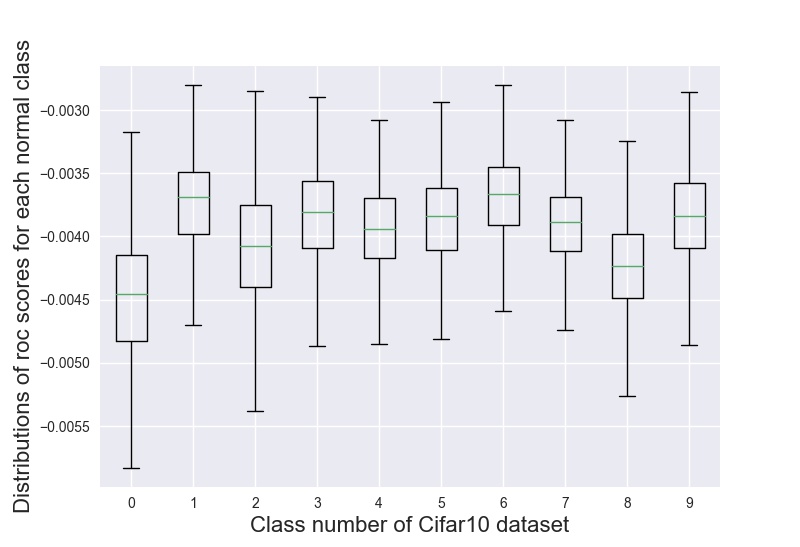}
\setlength{\belowcaptionskip}{-0.3cm}
\caption{ Anomaly score distributions on Cifar10 dataset\label{cifar-box}. The bottom and top of the box are first and third quartiles. Green lines in the box represent the median.}

\end{figure}
Figure \ref{cifar-roc-curve} shows the ROC curves for all experiments. Figure \ref{cifar-box} is a boxplots which represents the distribution of anomaly score for each class when class $0$ is normal data. The distributions of  anomaly scores for class 2, 8 overlap to class 0 to some degree. In our method, the ability to distinguish anomalies for each class is different.

\subsubsection{Experiment results on UCI datasets}
 We  select several small datasets  to show the performance of our methods. These datasets include KDDCUP99, cover type and shuttle. KDD dataset consists of five main classes. Only Normal class is normal data. All other classes are  anomalies. Shuttle dataset contains 9 attributes all of which are numerical. Approximately $80\%$ of the data belong to class 1. Cover type dataset predicts $7$ forest cover types from cartographic variables. Each vector consists of 54-dimensional attributes. For UCI experiments,  we set up normal data and anomalies as in Table \ref{UCI}. A small holdout set is used to decide the termination of each method.  We sample $80\%$ of normal data as training data. Other normal data and anomalies are used as test data.

\begin{table}[t]
\center\setlength{\belowcaptionskip}{5pt}
\caption{ Experimental set up on UCI datasets\label{UCI}}
\begin{tabular}{ccccccccccccc}
\hline
{Name} &{dataset} &{normal} & {anomalies}\\
\hline
KDD-A&KDD &normal&   attack  \\
\hline
COV-A&cover type&class 1,3,5,6,7&  2,4   \\
\hline
COV-B&cover type &class 2&    class 4 \\
\hline
SHU-A&shuttle &class 1&  class  2,3,4,5,6,7\\
\hline
\end{tabular}
\end{table}

%Before training model, we transfer the discrete attribute to numeric type and normalize the dataset.  All results are shown in Table \ref{UCIresult}.
%\begin{table}[t]
%\caption{Roc Scores on UCI Datasets \label{UCIresult}}
%\center
%\tiny
%\begin{tabular}{ccccccccc}
%%{p{1cm}|p{1cm}|p{1cm}|p{1cm}|p{1cm}|p{1cm}|p{1cm}}
%{Name}& {EMinLGAN-1}& {MinLGAN} & {GAN} & {IFOREST} & {OCSVM} & {VAE} & {AE}\\
%\hline
%KDD-A&-&-& 0.992& 0.991&0.982  & 0.995&0.937  \\
%\hline
%KDD-B&-&-& 0.991&0.997&  0.842& 0.823& 0.959 \\
%\hline
%COV-A&0.811&0.798& 0.793   &0.293 & 0.397& 0.743 &0.735  \\
%\hline
%COV-B&0.975&0.945& 0.931   &0.991& 0.997 & 0.956 &0.998  \\
%\hline
%SHU-A&0.987&0.985& 0.979   & 0.988& 0.947 & 0.802 &0.978\\
%\hline
%\end{tabular}
%\end{table}

\begin{table}[t]
\caption{ROC scores on UCI datasets \label{UCIresult}}
\setlength{\abovecaptionskip}{-10pt}
\center
\footnotesize
\begin{tabular}{ccccccccc}
%{p{1cm}|p{1cm}|p{1cm}|p{1cm}|p{1cm}|p{1cm}|p{1cm}}
\hline
{Name}& {KDD-A}& {COV-A} & {COV-B} & {SHU-A}\\
\hline
EMinLGAN-1&0.993  &0.811& 0.975&0.988 \\
\hline
MinLGAN&0.993  &0.798& 0.945&0.986 \\
\hline
GAN    &0.993 &0.793& 0.931&0.979  \\
\hline
IFOREST&0.991 &0.293& 0.991&0.988  \\
\hline
OCSVM  &0.982 &0.397& 0.997&0.947  \\
\hline
VAE    &0.995 &0.743& 0.956&0.802\\
\hline
AE     &0.937 &0.735& 0.998&0.978\\
\hline
\end{tabular}
\end{table}

From Table \ref{UCIresult},  we see that  our methods  have good performance on all experiments. On the KDD dataset, the GAN baseline performs as well as MinLGAN and EMinLGAN-1. OCSVM and IFOREST methods perform well except for COV-A. This is because normal data consists of several classes and there is not an effective feature extraction method. VAE and AE share the same network structure for the experiments. In our experiments, reconstruction probability based methods perform less stably than reconstruction error based methods for experiment SHU-A.
\section{CONCLUSION}

In this paper, we propose a  GAN-based  method  for anomaly detection. Our method demonstrates high performance on benchmark datasets, but is less stable compared with other methods because of uncertainty of anomalies trajectories and  training way of GAN. How to stabilize the performance of GAN  needs to be studied further in the future.

\vspace{0.5cm}
\begin{center}
ACKNOWLEDGEMENTS
\end{center}

This work has been supported by the National Natural Science Foundation of China (NSFC) grants 61721004 and 61773376.

\bibliographystyle{plain}
\bibliography{ijcai17}

\begin{thebibliography}{10}

\bibitem{ando2007clustering}
Shin Ando.
\newblock Clustering needles in a haystack: An information theoretic analysis
  of minority and outlier detection.
\newblock In {\em Seventh IEEE International Conference on Data Mining}, pages
  13--22. IEEE, 2007.

\bibitem{angelov2004approach}
Plamen Angelov.
\newblock An approach for fuzzy rule-base adaptation using on-line clustering.
\newblock {\em International Journal of Approximate Reasoning}, 35(3):275--289,
  2004.

\bibitem{augusteijn2002neural}
MF~Augusteijn and BA~Folkert.
\newblock Neural network classification and novelty detection.
\newblock {\em International Journal of Remote Sensing}, 23(14):2891--2902,
  2002.

\bibitem{campbell2001linear}
Colin Campbell and Kristin~P Bennett.
\newblock A linear programming approach to novelty detection.
\newblock In {\em Advances in neural information processing systems}, pages
  395--401, 2001.

\bibitem{chalapathy2017robust}
Raghavendra Chalapathy, Aditya~Krishna Menon, and Sanjay Chawla.
\newblock Robust, deep and inductive anomaly detection.
\newblock {\em arXiv preprint arXiv:1704.06743}, 2017.

\bibitem{chang2011libsvm}
Chih-Chung Chang and Chih-Jen Lin.
\newblock Libsvm: a library for support vector machines.
\newblock {\em ACM transactions on intelligent systems and technology},
  2(3):27, 2011.

\bibitem{chatzigiannakis2006hierarchical}
Vasilis Chatzigiannakis, Symeon Papavassiliou, Mary Grammatikou, and
  B~Maglaris.
\newblock Hierarchical anomaly detection in distributed large-scale sensor
  networks.
\newblock In {\em 11th IEEE Symposium on Computers and Communications}, pages
  761--767. IEEE, 2006.

\bibitem{clifton2006learning}
David~A Clifton, Peter~R Bannister, and Lionel Tarassenko.
\newblock Learning shape for jet engine novelty detection.
\newblock In {\em International Symposium on Neural Networks}, pages 828--835.
  Springer, 2006.

\bibitem{dai2017good}
Zihang Dai, Zhilin Yang, Fan Yang, William~W Cohen, and Ruslan Salakhutdinov.
\newblock Good semi-supervised learning that requires a bad gan.
\newblock {\em arXiv preprint arXiv:1705.09783}, 2017.

\bibitem{BDLZ2014}
Bo~Du and Liangpei Zhang.
\newblock A discriminative metric learning based anomaly detection method.
\newblock {\em IEEE Transactions on Geoscience and Remote Sensing}, 52(11):6844
  -- 6857, 2014.

\bibitem{DU2016115}
Bo~Du, Rui Zhao, Liangpei Zhang, and Lefei Zhang.
\newblock A spectral-spatial based local summation anomaly detection method for
  hyperspectral images.
\newblock {\em Signal Processing}, 124:115 -- 131, 2016.
\newblock Big Data Meets Multimedia Analytics.

\bibitem{eghbal2017likelihood}
Hamid Eghbal-zadeh and Gerhard Widmer.
\newblock Likelihood estimation for generative adversarial networks.
\newblock {\em arXiv preprint arXiv:1707.07530}, 2017.

\bibitem{erdogmus2004multivariate}
D~Erdogmus, R~Jenssen, YN~Rao, and JC~Principe.
\newblock Multivariate density estimation with optimal marginal parzen density
  estimation and gaussianization.
\newblock In {\em Machine Learning for Signal Processing. Proceedings of 14th
  IEEE Signal Processing Society Workshop}, pages 73--82. IEEE, 2004.

\bibitem{NIPS2014}
Ian Goodfellow, Jean Pouget-Abadie, Mehdi Mirza, Bing Xu, David Warde-Farley,
  Sherjil Ozair, Aaron Courville, and Yoshua Bengio.
\newblock Generative adversarial nets.
\newblock In Z.~Ghahramani, M.~Welling, C.~Cortes, N.~D. Lawrence, and K.~Q.
  Weinberger, editors, {\em Advances in Neural Information Processing Systems
  27}, pages 2672--2680. Curran Associates, Inc., 2014.

\bibitem{grover2017flow}
Aditya Grover, Manik Dhar, and Stefano Ermon.
\newblock Flow-gan: Bridging implicit and prescribed learning in generative
  models.
\newblock {\em arXiv preprint arXiv:1705.08868}, 2017.

\bibitem{heller2003one}
Katherine~A Heller, Krysta~M Svore, Angelos~D Keromytis, and Salvatore~J
  Stolfo.
\newblock One class support vector machines for detecting anomalous windows
  registry accesses.
\newblock In {\em Proc. of the workshop on Data Mining for Computer Security},
  volume~9, 2003.

\bibitem{jyothsna2011review}
V~Jyothsna, VV~Rama Prasad, and K~Munivara Prasad.
\newblock A review of anomaly based intrusion detection systems.
\newblock {\em International Journal of Computer Applications}, 28(7):26--35,
  2011.

\bibitem{larsen2015autoencoding}
Anders Boesen~Lindbo Larsen, S{\o}ren~Kaae S{\o}nderby, Hugo Larochelle, and
  Ole Winther.
\newblock Autoencoding beyond pixels using a learned similarity metric.
\newblock {\em arXiv preprint arXiv:1512.09300}, 2015.

\bibitem{liu2008isolation}
Fei~Tony Liu, Kai~Ming Ting, and Zhi-Hua Zhou.
\newblock Isolation forest.
\newblock In {\em Eighth IEEE International Conference on Data Mining}, pages
  413--422. IEEE, 2008.

\bibitem{mcbain2011feature}
Jordan McBain and Markus Timusk.
\newblock Feature extraction for novelty detection as applied to fault
  detection in machinery.
\newblock {\em Pattern Recognition Letters}, 32(7):1054--1061, 2011.

\bibitem{pokrajac2007incremental}
Dragoljub Pokrajac, Aleksandar Lazarevic, and Longin~Jan Latecki.
\newblock Incremental local outlier detection for data streams.
\newblock In {\em IEEE Symposium on Computational Intelligence and Data
  Mining}, pages 504--515. IEEE, 2007.

\bibitem{principi2017acoustic}
Emanuele Principi, Fabio Vesperini, Stefano Squartini, and Francesco Piazza.
\newblock Acoustic novelty detection with adversarial autoencoders.
\newblock In {\em International Joint Conference on Neural Networks}, pages
  3324--3330. IEEE, 2017.

\bibitem{ravanbakhsh2017abnormal}
Mahdyar Ravanbakhsh, Moin Nabi, Enver Sangineto, Lucio Marcenaro, Carlo
  Regazzoni, and Nicu Sebe.
\newblock Abnormal event detection in videos using generative adversarial nets.
\newblock {\em arXiv preprint arXiv:1708.09644}, 2017.

\bibitem{salimans2016}
Tim Salimans, Ian Goodfellow, Wojciech Zaremba, Vicki Cheung, Alec Radford, and
  Xi~Chen.
\newblock Improved techniques for training gans.
\newblock In {\em Advances in Neural Information Processing Systems}, pages
  2226--2234, 2016.

\bibitem{schlegl2017unsupervised}
Thomas Schlegl, Philipp Seeb{\"o}ck, Sebastian~M Waldstein, Ursula
  Schmidt-Erfurth, and Georg Langs.
\newblock Unsupervised anomaly detection with generative adversarial networks
  to guide marker discovery.
\newblock In {\em International Conference on Information Processing in Medical
  Imaging}, pages 146--157. Springer, 2017.

\bibitem{scholkopf2001}
Bernhard Sch{\"o}lkopf, John~C Platt, John Shawe-Taylor, Alex~J Smola, and
  Robert~C Williamson.
\newblock Estimating the support of a high-dimensional distribution.
\newblock {\em Neural computation}, 13(7):1443--1471, 2001.

\bibitem{schreyer2017detection}
Marco Schreyer, Timur Sattarov, Damian Borth, Andreas Dengel, and Bernd Reimer.
\newblock Detection of anomalies in large scale accounting data using deep
  autoencoder networks.
\newblock {\em arXiv preprint arXiv:1709.05254}, 2017.

\bibitem{stranjak2008multi}
Armin Stranjak, Partha~Sarathi Dutta, Mark Ebden, Alex Rogers, and Perukrishnen
  Vytelingum.
\newblock A multi-agent simulation system for prediction and scheduling of aero
  engine overhaul.
\newblock In {\em Proceedings of the 7th international joint conference on
  Autonomous agents and multiagent systems: industrial track}, pages 81--88.
  International Foundation for Autonomous Agents and Multiagent Systems, 2008.

\bibitem{yang2006support}
Qing Yang and Fangmin Li.
\newblock Support vector machine for intrusion detection based on lsi feature
  selection.
\newblock In {\em Sixth World Congress on Intelligent Control and Automation},
  volume~1, pages 4113--4117. IEEE, 2006.

\bibitem{yu2017open}
Yang Yu, Wei-Yang Qu, Nan Li, and Zimin Guo.
\newblock Open-category classification by adversarial sample generation.
\newblock {\em arXiv preprint arXiv:1705.08722}, 2017.

\end{thebibliography}

\end{document}